\documentclass[11pt]{article}
\usepackage[preprint]{acl}

\usepackage{times}
\usepackage{latexsym}

\usepackage[T1]{fontenc}

\usepackage[utf8]{inputenc}

\usepackage{microtype}

\usepackage{inconsolata}

\usepackage{amsmath}

\usepackage{multirow}
\usepackage{booktabs}

\usepackage{makecell}
\usepackage{tabu}

\usepackage{amssymb} 

\usepackage{array}

\usepackage{multirow}
\usepackage{caption}
\usepackage{subcaption}
\usepackage{amsfonts}
\usepackage{booktabs}
\usepackage{enumitem}
\usepackage{wrapfig}
\usepackage{listings}
\usepackage{framed}
\usepackage{tcolorbox}
\usepackage{CJKutf8}

\usepackage{thmtools, thm-restate}

\usepackage{soul}
\usepackage{hyperref}
\usepackage{url}
\definecolor{lightgray}{gray}{0.95}
\usepackage{avant}
\usepackage{pbox} 
\title{Shall Your Data Strategy Work? Perform a Swift Study}

\author{
MinLong Peng, Jingyi Yang, Zhongjun He, Hua Wu \\
NLP, Baidu Research, China\\
\texttt{pengminlong,yangjingyi01,hezhongjun,wu\_hua@baidu.com}
}

\begin{document}
\maketitle

\begin{CJK*}{UTF8}{gbsn}


\begin{abstract}

This work presents a swift method to assess the efficacy of particular types of instruction-tuning data, utilizing just a handful of probe examples and eliminating the need for model retraining. This method employs the idea of gradient-based data influence estimation, analyzing the gradient projections of probe examples from the chosen strategy onto evaluation examples to assess its advantages. Building upon this method, we conducted three swift studies to investigate the potential of Chain-of-thought (CoT) data, query clarification data, and response evaluation data in enhancing model generalization. Subsequently, we embarked on a validation study to corroborate the findings of these swift studies. 
In this validation study, we developed training datasets tailored to each studied strategy and compared model performance with and without the use of these datasets. The results of the validation study aligned with the findings of the swift studies, validating the efficacy of our proposed method.

\end{abstract}

\section{Introduction}

Instruction tuning has markedly improved the ability of large language models (LLMs) to adhere to human instructions \cite{openai2022openai,achiam2023gpt,zhou2024lima,sun2024principle}. Nonetheless, determining the best approaches for creating and utilizing diverse instruction-tuning data remains a persistent challenge  \cite{sprague2024cot,shen2024rethinking}. 
For instance, \textit{when confronted with a complex problem requiring multiple basic skills, should one focus on developing training data that mirrors the entire problem or first construct relevant training data around the foundational skills}?

To validate a particular data creation strategy, a substantial number of training examples are typically collected. The LLM is then retrained on this dataset and evaluated on a designated set \cite{liu2024what}. This process is often time-consuming and costly in terms of labor. Having a rapid method to assess the effectiveness of instruction-tuning data generated by a specific strategy would greatly benefit developers by enabling them to avoid wasting time on ineffective approaches.

In this work, we present a method to achieve this goal. This method is inspired by previous research on estimating the impact of individual training examples using gradient information \cite{pruthi2020estimating,xialess}. When confronted with a strategy awaiting validation, the method selects a few examples generated by the strategy as probes. It then computes the gradients of these probes at different states of the model. The strategy's effectiveness is determined by analyzing the gradient projection values between the probes and designed evaluation examples. For instance, to validate the strategy ``\textit{Enhance generalization on non-CoT data by incorporating CoT training data}", one could select a set of CoT examples as probes and non-CoT examples as evaluation examples. The method then calculates the gradient projection of probes onto the evaluation examples at different model states. Consistently positive and significant gradient projections suggest that the strategy is likely effective; otherwise, it may not be.

We apply our proposed method to three intuitive data creation strategies: incorporating \textit{CoT data}, \textit{query clarification data}, and \textit{response evaluation data} to enhance model generalization. These three types of data correspond to three essential capabilities: reasoning, problem analysis, and response evaluation, respectively. 
Our swift studies show that all three types of training data, particularly CoT training data, exhibit a greater capacity for model generalization compared to the general Question-Answer (QA) training data, albeit they may not perform as well for directly badcase resolving.

To validate the swift study and subsequently affirm the effectiveness of our proposed method, we have amassed a substantial amount of training data for each data creation strategy. By comparing the performance of LLMs trained with and without this data, we can assess the efficacy of the data creation strategies. The results obtained from this validation study are in harmony with those from the swift studies, thereby confirming the veracity of the swift studies and validating our proposed method. 

\section{Related Work}
\subsection{Instruction Tuning and Data Creation Strategies}

LLM alignments usually encompass instruction tuning and reinforcement learning with human feedback (RLHF) \cite{ouyang2022training}. Instruction tuning, often referred to as supervised fine-tuning (SFT),  trains LLMs on a dataset consisting of (QUERY, RESPONSE) pairs in a supervised fashion. This effectively bridges the gap between the next-word prediction pertaining objective of LLMs and
the users’ objective of having LLMs adhere to human instructions \cite{zhang2023instruction}.

In recent years, many heuristic strategies have been explored for creating instruction-tuning data \cite{sprague2024cot}. \citet{wang2023self} proposed to generate instruction tuning data using ChatGPT, prompting ChatGPT to generate both the QUERY and RESPONSE. In the following, quite a few strategies have been proposed to improve the quality of instruction-tuning data. \citet{wang2023noise,weber2024donkii} proposed to remove low-quality data from instruction tuning data with the consideration that low-quality data may mislead LLMs. In another direction, some work found that increasing data quality and diversity instead of quantity can effectively induce instruction following abilities \cite{zhou2024lima,liumakes,xialess}. Thereby, they proposed different strategies to create, or select from existing datasets, high-quality and diverse instruction tuning data. 
For instance, \citet{shen2024rethinking} found out through experimental studies that long responses are surprisingly more effective for instruction tuning. 
More recently, \citet{ding2024mitigating} proposed an innovative Socratic-guided sampling strategy to select instruction-tuning data, improving the sampling efficiency for challenging heavy-tailed data.

However, there lack of a principled understanding of what makes good instruction tuning data for alignment. The building of a data creation strategy predominantly relies on developers' instruction, followed by extensive data creation and experimental validation.
This process is often time-consuming and labor-cost. In this work, we design a swift method to study a strategy's effectiveness using only a handful of examples. This can greatly reduce the cost of strategy validation. In addition, this method may be combined with existing data sampling strategies, e.g., that proposed by \citet{ding2024mitigating}, to achieve further improvement. We leave this topic for future work.

\subsection{Influence Estimation by Gradient Descent}

Data influence estimation using gradient descent has been used in many tasks, such as identifying mislabeled examples \cite{pruthi2020estimating}, analyzing memorization effects \cite{feldman2020neural}, and deriving various interpretability insights \cite{madsen2022post}. In the LLM setting, this kind of method was mainly used for training data selection. For instance, \cite{han2023understanding} used model gradients to select a subset of pretraining data
that supports ICL and accordingly analyze the characteristics of the supportive pertaining data. \cite{engstromdsdm,xialess} traced the projections of training examples' gradients onto those of evaluation examples, and accordingly selected the most effective training examples for evaluation performance. In this work, we employ the idea of gradient-based influence estimation to study the effectiveness of particular data creation strategies on target tasks efficiently. 

\section{Efficacy Study of Instruction-Tuning Data using Gradient Descent}

\textbf{Train-evaluation influence estimation with gradient descent.} Given a training example $z$ and an evaluation example $z_0$, \citet{pruthi2020estimating} defines the influence of $z$ to $z_0$ during the whole model training process when the batch size is 1 as follows:
\begin{equation}
    \text{TracInIdeal}(z, z_0) = \sum_{t: z_t = z} \ell(z_0;\theta_t) - \ell(z_0; \theta_{t+1}),
\end{equation}
where $t$ denotes the model training step, $\theta_t$ denotes the model state at step $t$, and $\ell(z;\theta)$ denotes the loss of $z$ at $\theta_t$. Here, $\ell(z_0;\theta_t) - \ell(z_0; \theta_{t+1})$ is often referred to as step influence of $z$ to $z_0$. 

With the first-order Taylor expansion of the loss, the loss of $z_0$ can be approximated by:
\begin{equation*}
    \ell(z_0; \theta_{t+1}) \approx \ell(z_0; \theta_t) + \langle \nabla \ell(z_0; \theta_t), \theta_{t+1} - \theta_t \rangle.
\end{equation*}
For ease of exposition, assume the model is trained with SGD with batch size 1 and learning rate $\eta_t$. It can write the SGD update as $\theta_{t+1} - \theta_t = -\eta_t \nabla \ell(z; \theta_t)$. Then, the influence of the training example $z$ to the evaluation example $z_0$ can be approximated by:
\begin{equation}
\begin{split}
     \text{TracInIdeal}(z, z_0) \approx \\ -\sum_{t: z_t = z} \eta_t \langle \nabla \ell(z; \theta_t), \nabla \ell(z_0; \theta_t) \rangle.
\end{split}
\end{equation}

\textbf{Swift data-creation strategy study with gradient descent.} While \citet{pruthi2020estimating} used the insight to identify mislabeled training data and \citet{xialess} applied a similar formula to design a data selection strategy, we instead used the idea to validate the effectiveness of instruction-tuning data creation strategies. 
Two critical aspects render the above algorithm unsuitable for direct application in our setting. 
Firstly, it requires training the model from scratch with the generated examples by the investigated strategy. This is often time-consuming and resource-intensive. 
Secondly, step influences in different model states are not comparable as the loss generally decreases as model training. With this consideration, we introduce a relative influence score of $z$ on $z_0$ at $\theta_t$ defined as follows:
\begin{equation}
    \text{RelInf}(z, z_0;\theta_t) = \frac{\langle \nabla \ell(z; \theta_t), \nabla \ell(z_0; \theta_t) \rangle}{\langle \nabla \ell(z_0; \theta_t), \nabla \ell(z_0; \theta_t) \rangle}.
\end{equation}
$\text{RelInf}(z, z_0;\theta_t)$ measures the degree to which training the model on $z$ reduces the loss on $z_0$ relative to training directly on $z_0$ at a given model state $\theta_t$.
It is comparable at different states of $\theta_t$. 
Notably, $z$ does not need to be included in the model training process, obviating the need to retrain the model for each strategy under consideration. 
We can trace this value to analyze the influence of $z$ on $z_0$ over different model states and gauge the ultimate influence of $z$ to $z_0$. For instance, if $\text{RelInf}(z, z_0;\theta_t)$ remains high (e.g., close to 1) across various model states, it suggests that training with $z$ yields a comparable effect on $z_0$ as training with $z_0$ itself, typically leading to a minimal loss on $z_0$.

In the subsequent section, we show how to efficiently utilize $\text{RelInf}(z, z_0; \theta_t)$ to investigate the effectiveness of three intuitive data creation strategies for instruction-tuning. In these studies, we calculate the gradient of a sequence-to-sequence instruction-tuning example $z$ by summing the gradients of all tokens in its completion.

\section{Swift Study on Three Strategies}

We employ our methodology to investigate the efficacy of \textit{chain-of-thought training data, query clarification training data}, and \textit{response evaluation data}, which correspond to three basic capabilities: reasoning, problem analysis, and response evaluation, respectively. Table \ref{tab:probe_case} in the Appendix shows demo examples generated using these three strategies.

\textbf{Evaluation.}
In these studies, we meticulously investigate the impact of the studied strategy in the \textbf{in-task} context and the \textbf{cross-task} context, respectively. Let $z'_i$ denote an example created from a general QA example $z_i$ according to the studied strategy. 
In the in-task context, we evaluate the effect of $z'_i$ on $z_i$ or its analogous instances, shedding light on the strategy's capacity for badcase resolving (suppose $z_i$ is the badcase we aim to address). 
In the cross-task context, we explore the influence of $z'_i$ on QA examples drawn from different tasks, $z_{j\neq i}$, thereby unveiling the strategy's capacity for cross-task generalization. 
Formally, let $\mathcal{Z} = (z_1, \cdots, z_n)$ denote the set of evaluation examples from distinct tasks, and $\mathcal{Z}' = (z'_1, \cdots, z'_n)$ denote a corresponding data set of $\mathcal{Z}$, where $z'_i$ is closely related to $z_i$, e.g., $z'_i$ is built from $z_i$ or one of its analogus examples using the studied strategy. We introduce an average in-task influence score and an average cross-task influence score to evaluate the influence of $\mathcal{Z}'$ on $\mathcal{Z}$ in the in-task and the cross-task context, respectively:
\begin{align}
    s_{\text{in}}(\mathcal{Z}', \mathcal{Z}, t)  &= \frac{1}{n} \sum_{i=1}^n \text{RelInf}(z'_i, z_i; \theta_t), \label{eq:in-task} \\ 
    s_{\text{cross}}(\mathcal{Z}', \mathcal{Z}, t) &= \frac{\sum_{i=1}^n \sum_{j\neq i} \text{RelInf}(z'_i, z_j; \theta_t)}{n(n-1)}.\label{eq:cross-task} 
\end{align}

\textbf{Setup.} We conducted swift studies utilizing Chinese-LLaMA2-7B \cite{cui2023efficient}, which extends the Chinese vocabulary of LLaMA2 \cite{touvron2023llama} and undergoes incremental pretraining on mixed corpora. To obtain model stages with varying levels of instruction-following proficiency, we gathered 6,400 examples for instruction-tuning. We trained Chinese-LLaMA2-7B 6 epochs from scratch using these examples, saving a checkpoint every 50 steps with a global batch size set to 32. The learning rate was initially set to 1e-5 and linearly decayed to 1e-7 with 10 warmup steps.
To enhance efficiency, we incorporated LoRA \cite{hu2021lora} to minimize the number of trainable parameters and expedite influence calculations. Specifically, we froze the pre-trained weights and introduced a low-rank adapter to the linear layers of the network's down and up projections. The rank of LoRA was set to 64, and the alpha hyper-parameter was adjusted to 512. For clarity, we will refer to the 6,400 instruction-tuning examples as the \textbf{base dataset}, and the finetuned model obtained at the final step as the \textbf{base model}. It is important to note that all probe examples collected in subsequent studies are excluded from the base dataset.

\subsection{Can CoT Training Data Improve Model Generalization on non-CoT Queries?}

\begin{figure}[t]
    \vspace{-10pt}
    \centering
    \includegraphics[width=\linewidth]{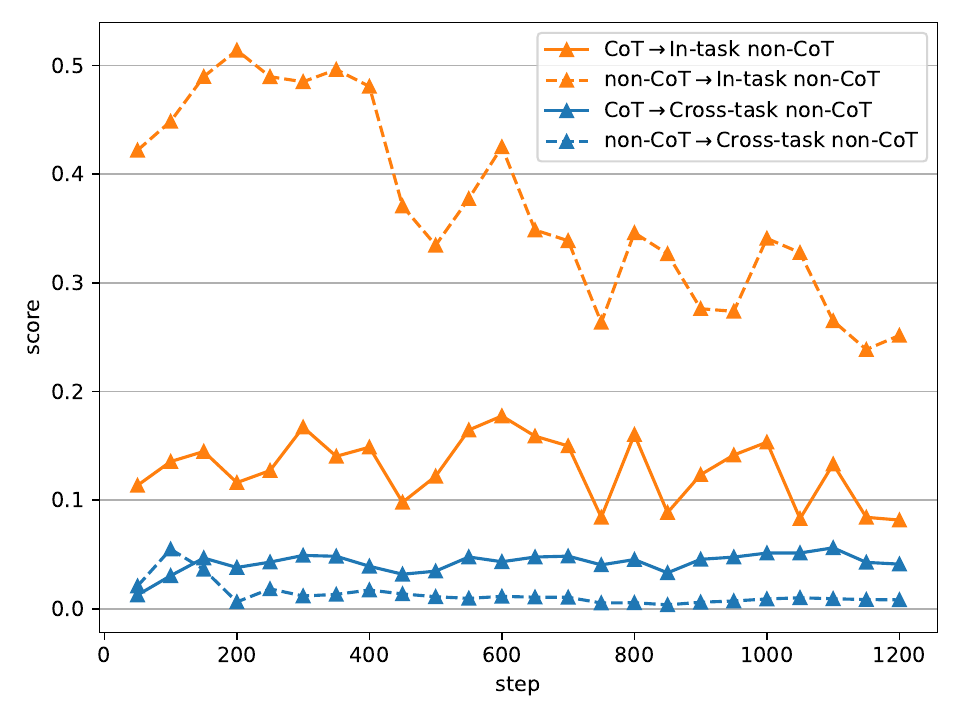}
    \caption{Swift study on CoT data}
    \vspace{-10pt}
    \label{fig:cot-influence}
\end{figure}

Chain-of-Though (CoT) is a prevalent approach to improving the performance of LLMs, in which the LLM generates explicit intermediate reasoning steps before arriving at a final answer \cite{wei2022chain}. 
However, the explicit reasoning process inherent in CoT can be computationally demanding. Numerous users, particularly those charged for LLM services based on token usage, frequently prefer direct answers without the intervening reasoning steps. This preference raises an intriguing question: \textit{In scenarios where users only require non-CoT responses, is it necessary to introduce CoT training data?}

To address this inquiry, we conducted a swift study using our proposed method. For this investigation, we collected several non-CoT examples from disparate tasks, denoted as $\mathcal{Z}$. Table \ref{tab:reference_case} in the Appendix shows some of the examples. Based on each example $z_i \in \mathcal{Z}$, we prompt Yiyan \cite{wenxinyiyan} to generate a similar non-CoT example $z'_i$, obtaining another non-CoT data set $\mathcal{Z}'$. Additionally, based $z_i$, we create a CoT example, $z''_i$, by appending an additional command to the query of $z_i$, asking for a reason before the conclusion, and adding some reasons to the response of $z_i$. We denote the resulting CoT data set as $\mathcal{Z}''$.   

To assess the effectiveness of the strategy in the in-task context, we track the value of $s_{\text{in}}(\mathcal{Z}'', \mathcal{Z}', t)$. For comparative purposes, we report the value of $s_{\text{in}}(\mathcal{Z}, \mathcal{Z}', t)$. 
For brevity and based on the meanings of these metrics, we refer to the two metrics as $\textbf{CoT}\rightarrow\textbf{In-task non-CoT}$ and $\textbf{non-CoT}\rightarrow\textbf{In-task non-CoT}$, respectively.
To assess the strategy's effectiveness in the cross-task context, we monitor the value of $s_{\text{cross}}(\mathcal{Z}'', \mathcal{Z}', t)$. For comparative purposes, we report the value of $s_{\text{cross}}(\mathcal{Z}, \mathcal{Z}', t)$. Based on their meaning, we abbreviate the two metrics as $\textbf{CoT}\rightarrow\textbf{Cross-task non-CoT}$ and $\textbf{non-CoT}\rightarrow\textbf{Cross-task non-CoT}$, respectively.

\begin{figure*}[t]
    \centering
    \begin{subfigure}[b]{0.47\textwidth}
        \includegraphics[width=\textwidth]{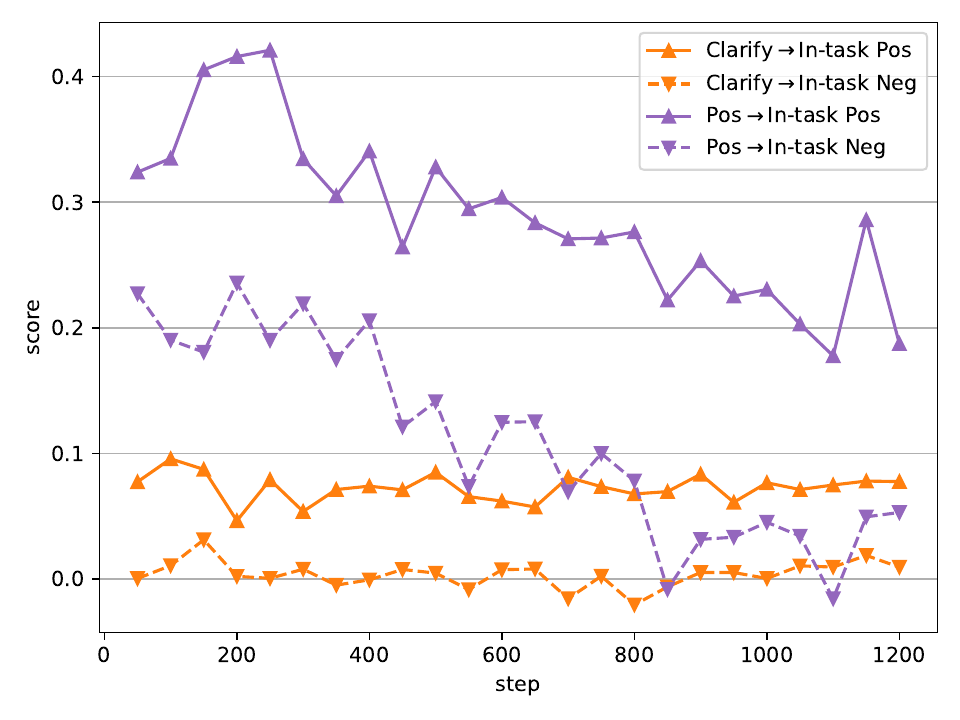}
        \caption{In the in-task context}
        \label{fig:query-clarification-in-task}
    \end{subfigure}
    \hfill
    \begin{subfigure}[b]{0.47\textwidth}
        \includegraphics[width=\textwidth]{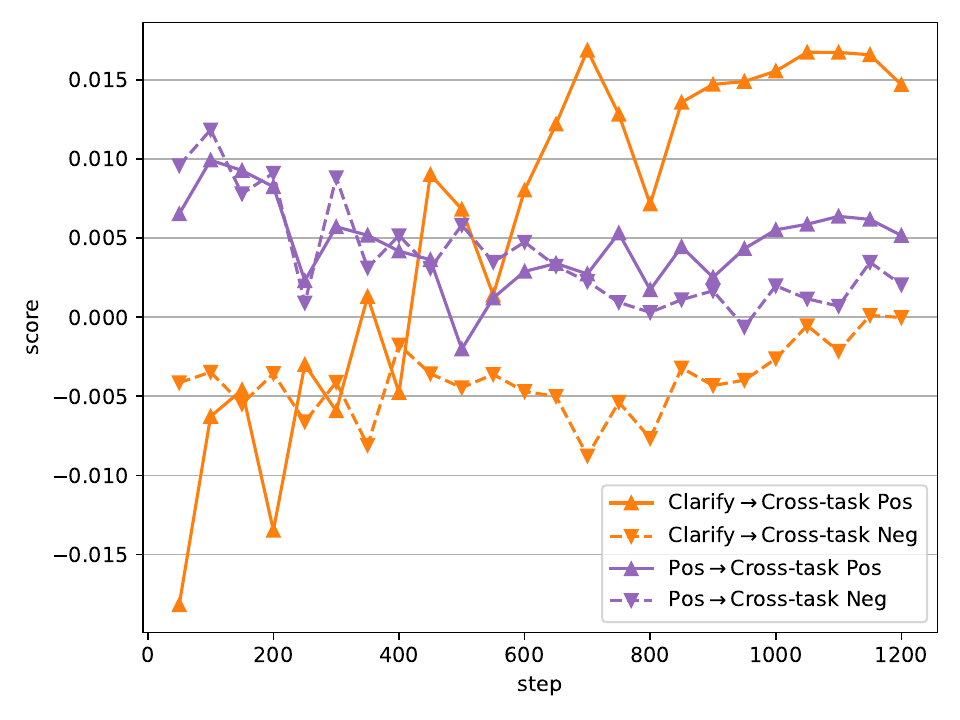}
        \caption{In the cross-task context}
        \label{fig:query-clarification-cross-task}
    \end{subfigure}
    \caption{Swift study on query clarification data}
    \label{fig:query-clarification-influence}
\end{figure*}

Figure \ref{fig:cot-influence} depicts the values of these metrics by step, allowing us to derive several key observations:
\textbf{1).} ${\text{CoT} \rightarrow \text{In-task non-CoT}}$ exhibits consistently high positive values across all steps. However, in comparison to ${\text{non-CoT} \rightarrow \text{In-task non-CoT}}$, its value is still notably lower. This implies that while CoT training data can assist in addressing badcases, it is less efficient compared to its non-CoT counterpart.
\textbf{2).} ${\text{non-CoT} \rightarrow \text{Cross-task non-CoT}}$ maintains a low value (slightly above zero) across all steps. This indicates the limited cross-task generalization ability of non-CoT training data. It may also explain why diversity of training data is more crucial than data size for LLM alignment \cite{liu2024what,zhou2024lima}.
\textbf{3).} After about 150 steps, ${\text{CoT} \rightarrow \text{Cross-task non-CoT}}$ consistently exceeds ${\text{non-CoT} \rightarrow \text{Cross-task non-CoT}}$ in value and remains about 0.05. This suggests that CoT data possesses superior cross-task generalization capabilities compared to non-CoT data. The introduction of CoT data, therefore, appears to be a promising strategy for enhancing cross-task generalization. Moreover, ${\text{CoT} \rightarrow \text{Cross-task non-CoT}}$ does not decrease but tends to increase by step, contrasting with the in-task metrics. We hypothesize that the cross-task generalization of CoT examples to non-CoT examples relies more on the inherent knowledge of LLMs rather than the instruction data itself. Further investigation into this phenomenon is left for future work.

\begin{tcolorbox}[colback=lightgray, colframe=black, boxrule=0.8pt, rounded corners, fontupper=\normalfont] \label{sum:cot}
\textbf{Swift Study Conclusion}: 
While CoT training data might not be as effective as non-CoT data in handling non-CoT badcases within similar tasks, it exhibits superior generalization capabilities on data from disparate tasks.

\end{tcolorbox}

\subsection{Can Query Clarification Training Data Improve Model Generalization on Confused Queries}

In the realm of LLMs, the quality of prompts is pivotal \cite{ekin2023prompt,white2023prompt}. Frequently, a meticulous refinement of the prompt can elevate the model's satisfaction rate from 0\% to a perfect 100\%. Yet, in practical scenarios, the majority of users do not invest significant time in crafting prompts or they do not know how to do that. Their prompts are frequently simplistic, insufficient, and occasionally perplexing. Revising these prompts with more accurate and detailed instructions typically yields a marked enhancement in the model's efficacy. This leads us to ponder: \textit{Can LLMs learn from query clarification training data to correctly understand users' simplistic and insufficient queries?} 

\begin{figure*}[t]
    \centering
    \begin{subfigure}[b]{0.47\textwidth}
        \includegraphics[width=\textwidth]{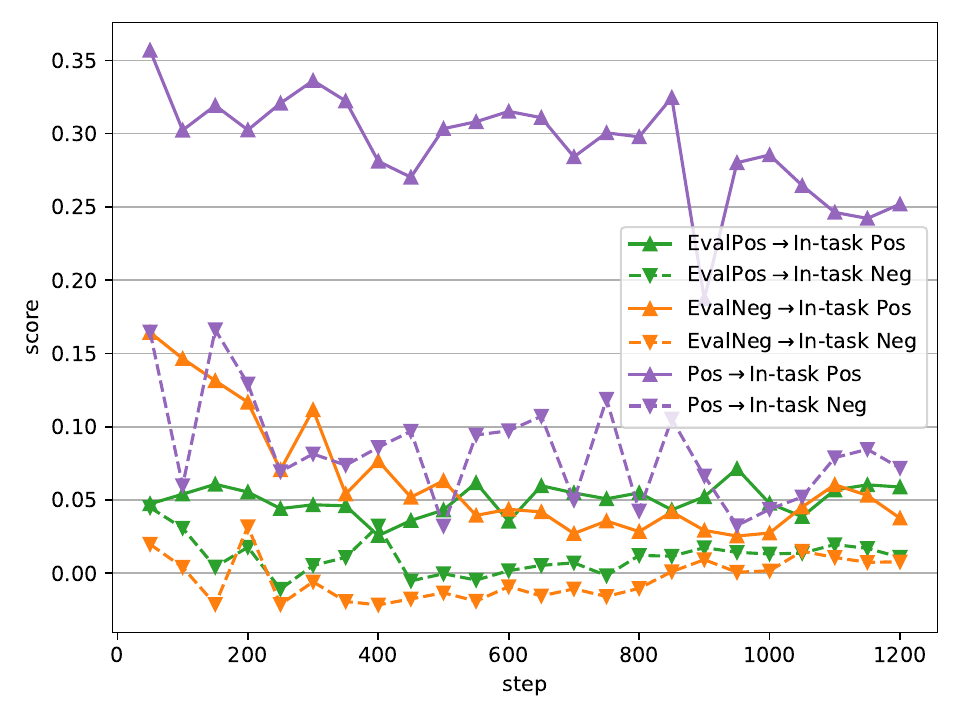}
        \caption{In the in-task context}
        \label{fig:eval-in-task}
    \end{subfigure}
    \hfill
    \begin{subfigure}[b]{0.47\textwidth}
        \includegraphics[width=\textwidth]{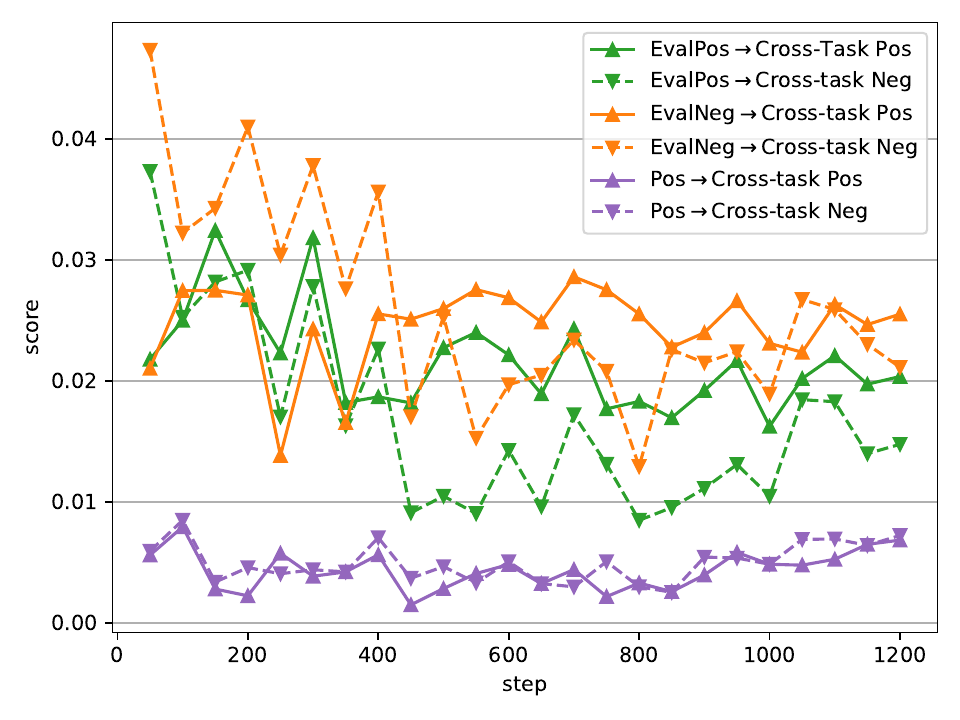}
        \caption{In the cross-task context}
        \label{fig:eval-cross-task}
    \end{subfigure}
    \caption{Swift study on response evaluation data}
    \label{fig:eval-influence}
\end{figure*}

To tackle this question, we perform a study employing our methodology. For this study, we gather 10 examples from disparate tasks, $\mathcal{Z}_{neg} = \{z_{n1}, \cdots, z_{nm}\}$, whose responses are generated by the base model and deemed unsatisfactory. 
Subsequently, we revise the query of each negative example so that the base model can generate a satisfactory response. We create a positive data set, $\mathcal{Z}_{pos}$, using the revised queries and their corresponding responses, and a query clarification data set, $\mathcal{Z}_{clarify}$, using the original misbehaved queries and their corresponding revised queries. After that, we generate a negative data set $\mathcal{Z}'_{neg}$ by prompting Yiyan \cite{wenxinyiyan} to generate a similar example for each example in $\mathcal{Z}_{neg}$. Similary, we generate a positive data set $\mathcal{Z}'_{pos}$ from $\mathcal{Z}_{pos}$. 

To evaluate the influence of the query clarification training data in the in-task context, we track the average in-task influence score of $\mathcal{Z}_{clarify}$ on $\mathcal{Z}'_{pos}$ and $\mathcal{Z}'_{neg}$, respectively. We abbreviate the two metrics as ${\textbf{Clarify} \rightarrow \textbf{In-task Pos}}$ and ${\textbf{Clarify} \rightarrow \textbf{In-task Neg}}$, respectively. 
For comparison, we report the average in-task influence score of $\mathcal{Z}_{pos}$ on $\mathcal{Z}'_{pos}$ and $\mathcal{Z}'_{neg}$, respectively. We abbreviate the two metrics as ${\textbf{Pos} \rightarrow \textbf{In-task Pos}}$ and ${\textbf{Pos} \rightarrow \textbf{In-task Neg}}$, respectively. As for the cross-task context, we track the average cross-task influence score of $\mathcal{Z}_{clarify}$ on $\mathcal{Z}'_{pos}$ and $\mathcal{Z}'_{neg}$, respectively. The two metrics are abbreviated as ${\textbf{Clarify} \rightarrow \textbf{Cross-task Pos}}$ and ${\textbf{Clarify} \rightarrow \textbf{Cross-task Neg}}$, respectively.
For comparison, we report the average cross-task influence score of $\mathcal{Z}_{pos}$ on $\mathcal{Z}'_{pos}$ and $\mathcal{Z}'_{neg}$, respectively, with the resulting metrics being abbreviated as ${\textbf{Pos} \rightarrow \textbf{Cross-task Pos}}$ and ${\textbf{Pos}\rightarrow \textbf{Cross-task Neg}}$, respectively.

Figure \ref{fig:query-clarification-in-task} and \ref{fig:query-clarification-cross-task} illustrate the results in the in-task and cross-task contexts, respectively. From the two figures, we can observe: 
\textbf{1).} ${\text{Clarify}\rightarrow \text{In-task Pos}}$ remains positive across steps and notably higher than  ${\text{Clarify} \rightarrow \text{In-task Neg}}$. This indicates that query clarification data can aid the model in achieving better in-task generalization. However, its value is still notably smaller than ${\text{Pos}\rightarrow \text{Corr. Pos}}$ across steps. This points towards the relative inefficiency of query clarification data in comparison to general QA data for in-task generalization. Nevertheless, it is noteworthy that ${\text{Clarify}\rightarrow \text{In-task Pos}}$ exhibits remarkable stability over step, showing only a slight decrease during the initial phases. On the other hand, ${\text{Pos}\rightarrow \text{In-task. Pos}}$ experiences a noticeable decline by step. This contrast suggests that query clarification data holds promise in resisting overfitting. 
\textbf{2).} In the cross-task setting, ${\text{Clarify}\rightarrow \text{Cross-task Pos}}$ begins with a value less than zero, along with ${\text{Pos}\rightarrow \text{Cross-task Pos}}$. Over time, this value progressively increases, ultimately surpassing zero and ${\text{Pos}\rightarrow \text{Cross-task Pos}}$. This trend indicates that clarifying queries positively impacts cross-task generalization. Furthermore, it is advisable to incorporate query clarification data at the later stages of model learning. Of course, the maximum value of ${\text{Clarify}\rightarrow \text{Cross-task Pos}}$ reaches only 0.015, and ${\text{Clarify}\rightarrow \text{Cross-task Neg}}$ remains near zero. This implies that a substantial number of query clarification examples is required to achieve a noticeable improvement. Fortunately, as we will demonstrate in the following section, generating a substantial amount of query clarification data can be straightforward and quite efficient.
    
\begin{tcolorbox}[colback=lightgray, colframe=black, boxrule=0.8pt, rounded corners, fontupper=\normalfont] \label{sum:cot}
\textbf{Swift Study Conclusion}: 
Similar to CoT training data, query clarification data might be not so effective as general QA training data in handling badcases, it exhibits superior cross-task generalization capabilities. In addition, it is more effective when introduced during the later stages of model training.

\end{tcolorbox}

\subsection{Can Response Evaluation Training Data Improve Model Generalization}

Lastly, we investigate \textit{whether it can improve models' performance on general QA tasks by enhancing the models' response evaluation capability.}
The rationale behind this inquiry stems from the belief that response evaluation capability equips the model with the insight to discern between superior and inferior responses, potentially leading to improved performance on general QAs. If this hypothesis holds, we can harness the vast amount of open- and closed-source response evaluation data, or we can automatically generate evaluation data based on the manual labels (e.g., the response is good or bad) to our fullest advantage.

To carry out the study, we reused the previously collected positive and negative examples $(\mathcal{Z}_{pos}, \mathcal{Z}_{neg}, \mathcal{Z}'_{pos}, \mathcal{Z}'_{neg})$ from the above query clarification study. Following this, we instructed Yiyan \cite{wenxinyiyan} to automatically assess the response quality of each example, providing comprehensive evaluation results for each instance. Based on these evaluations, we constructed corresponding evaluation data sets, denoted as $(\mathcal{E}_{pos}, \mathcal{E}_{neg}, \mathcal{E}'_{pos}, \mathcal{E}'_{neg})$. Herein, we abbreviate $\mathcal{E}_{pos}$ and $\mathcal{E}'_{pos}$ as EvalPos, and abbreviate $\mathcal{E}_{neg}$ and $\mathcal{E}'_{neg}$ as EvalNeg. We trace the average in-task influence score of $\mathcal{E}_{pos}$ on $\mathcal{Z}'_{pos}$ and $\mathcal{Z}'_{neg}$, respectively, and trace that of $\mathcal{E}_{neg}$ on $\mathcal{Z}'_{pos}$, and $\mathcal{Z}'_{neg}$, respectively. We abbreviate the four metrics as $\textbf{EvalPos}\rightarrow\textbf{In-task Pos}$, $\textbf{EvalPos}\rightarrow\textbf{In-task Neg}$, $\textbf{EvalNeg}\rightarrow\textbf{In-task Pos}$, and $\textbf{EvalNeg}\rightarrow\textbf{In-task Neg}$, respectively.
Similarly, we report the average in-task influence scores of $\mathcal{Z}_{pos}$ on $\mathcal{Z}'_{pos}$ and $\mathcal{Z}'_{neg}$, respectively, for comparison. As for the cross-task context, we report the average cross-task influence score of $\mathcal{E}_{pos}$ on $\mathcal{Z}'_{pos}$ and $\mathcal{Z}'_{neg}$, respectively, and that of $\mathcal{E}_{neg}$ on $\mathcal{Z}'_{pos}$ and $\mathcal{Z}'_{neg}$, respectively. We abbreviate the four metrics as $\textbf{EvalPos}\rightarrow\textbf{Cross-task Pos}$, $\textbf{EvalPos}\rightarrow\textbf{Cross-task Neg}$, $\textbf{EvalNeg}\rightarrow\textbf{Cross-task Pos}$, and $\textbf{EvalNeg}\rightarrow\textbf{Cross-task Neg}$, respectively. For comparative purposes, we also report the average cross-task influence score of $\mathcal{Z}_{pos}$ on $\mathcal{Z}'_{pos}$ and $\mathcal{Z}'_{neg}$, respectively, abbreviating as $\textbf{Pos}\rightarrow\textbf{Cross-task Pos}$ and $\textbf{Pos}\rightarrow\textbf{Cross-task Neg}$.

Figure \ref{fig:eval-in-task} and \ref{fig:eval-cross-task} depict the results of this study. From the figures, we can observe:
\textbf{1).} In the in-task setting, ${\text{EvalPos} \rightarrow \text{In-task Pos}}$ and  ${\text{EvalNeg} \rightarrow \text{In-task Pos}}$ remain positive (about 0.05) and keep larger than ${\text{EvalPos} \rightarrow \text{In-task Neg}}$ and ${\text{EvalNeg} \rightarrow \text{In-task Neg}}$. This suggests that evaluation data can improve model performance in the in-task context. Meanwhile, their values are much smaller than ${\text{Pos} \rightarrow \text{In-task Pos}}$, suggesting the inferior effectiveness of response evaluation data for in-task generalization compared to the ordinary QA data.
\textbf{2).} In the cross-task setting, ${\text{EvalPos} \rightarrow \text{Cross-task Pos}}$ and ${\text{EvalNeg} \rightarrow \text{Cross-task Pos}}$ maintains larger values than ${\text{Pos} \rightarrow \text{Cross-task Pos}}$. In addition, their values are generally larger than those of ${\text{EvalPos} \rightarrow \text{Cross-task Neg}}$ and ${\text{EvalNeg} \rightarrow \text{Cross-task Neg}}$ at the later stages of model training. This suggests a good cross-task generalization ability of response evaluation data.


\begin{tcolorbox}[colback=lightgray, colframe=black, boxrule=0.8pt, rounded corners, fontupper=\normalfont] \label{sum:cot}
\textbf{Swift Study Conclusion}: 
Similar to CoT and query clarification training data, response evaluation data might be not so effective as general QA training data in handling badcases, it exhibits superior cross-task generalization capabilities. 
\end{tcolorbox}

\section{Swift Study Validation}
To verify the correctness of the above swift studies and confirm the validity of our proposed methodology, we have designed a validation study wherein the model is retrained using training data generated by the strategies under investigation.

\subsection{Setup}
The validation study was conducted on Chinese-LLaMA2-7B and Chinese-LLaMA2-13B \cite{cui2023efficient} models with full-parameter training. Without confusion, we refer to the base model in the following as the model trained six epochs using the base dataset. For evaluation, we gathered 402 queries from real users, designated as $\mathcal{Q}_{eval}$. Most $\mathcal{Q}_{eval}$ require responses in a non-CoT mode. For efficient evaluation, we asked annotators to write a golden response and key checkpoints to check for each query of $\mathcal{Q}_{eval}$. Then, we design an automatic evaluation system using GPT-4o (2024-08-06 version) based on the golden response and checkpoints to evaluate whether a response fully meets the requirements of the query (the label is only 0 or 1) and accordingly report accuracy on $\mathcal{Q}_{eval}$. The consistency rate between human evaluation and system evaluation reaches about 87\%.

To validate the swift studies in the in-task context, we utilized Yiyan \cite{wenxinyiyan} to generate a similar example for each one in $\mathcal{Q}_{eval}$, obtaining a positive training data set, $\mathcal{D}_{pos}$. Based on $\mathcal{D}_{pos}$, we created a CoT training data set $\mathcal{D}_{cot}$, a query clarification data set $\mathcal{D}_{clarify}$, and a response evaluation data set $\mathcal{D}_{eval}$. We fine-tune the base model three epochs using each of the obtained data sets and evaluate the resulting models' performance on $\mathcal{Q}_{eval}$. In these experiments, the learning rate was set to $1e^{-6}$ and the batch size was set to 32. To assess the swift studies in the cross-task context, we generated a CoT training data set, a query clarification data set, and a response evaluation dataset from the base dataset. We then fine-tuned the pre-trained model using combinations of these obtained datasets. In this setting, the learning rate was initialized to $1e^{-5}$ and linearly decayed to $1e^{-7}$. The batch size was set to 32.  

\textbf{Data Generation.} 
Given a reference example, we supply Yiyan with the reference query and response, instructing it to generate a response with a detailed reasoning process to generate the CoT example. A similar process is applied to generate the response evaluation example but with a different prompt. As for generating the query clarification example, manually revising the reference query is time-consuming and impractical. Inspired by the observation that appending the query's requirements to its end often enhances the model performance, we utilized Yiyan to automatically extract the requirements of the reference query. The extracted requirements were then appended to the reference query, creating a clarified query. The pairing of the reference query with the clarified counterpart formed the query clarification example.

\subsection{Results}


\textbf{In the In-task Context.} Table \ref{tab:in-task-verify} presents the performance of the base model fine-tuned using different data sets created from $\mathcal{D}_{pos}$. The table indicates that the datasets contribute to performance improvements in the following order: $\mathcal{D}_{pos} > \mathcal{D}_{cot} > \mathcal{D}_{eval} > \mathcal{D}_{clarify} > 0$. The data generated by the three strategies under investigation can all enhance model performance on the evaluation data set but falls short compared to $\mathcal{D}_{pos}$. 
This corroborates our findings from the corresponding swift studies conducted in the in-task context.

\begin{table}[t]
     \centering
     \resizebox{\columnwidth}{!}{
     \begin{tabular}{l|c|c}
     \hline
         Training Data & CLLaMA2-7B & CLLaMA2-13B \\ \hline \hline
          Base                              & 52.74\%   & 65.42\% \\  
          \quad $\rightarrow$ $\mathcal{D}_{pos}$       & 59.70\%   & 70.40\% \\
          \quad $\rightarrow$ $\mathcal{D}_{cot}$       & 55.22\%   & 68.40\%\\ 
          \quad $\rightarrow$ $\mathcal{D}_{clarify}$   & 52.98\%   & 66.17\% \\ 
          \quad $\rightarrow$ $\mathcal{D}_{eval}$      & 54.48\%   & 67.67\% \\ \hline
    \end{tabular}
    }
     \caption{Model performance (Accuracy) in the in-task context when using different datasets derived from $\mathcal{D}_{eval}$ to continue fine-tuning the base model. CLLaMA2-7B denotes the Chinese-LLaMA2-7B model.}
     \label{tab:in-task-verify}
    \vspace{-10pt}
 
\end{table}


\textbf{In the Cross-task Context.} Table \ref{tab:cross-task-verify} displays the performance of models fine-tuned using various data sets built from the base dataset. From this table, we can observe:
\textbf{1).} Models fine-tuned on Base + CoT, Base + Clarify, and Base + Eval demonstrate enhanced performance compared to their counterparts fine-tuned solely on Base. Notably, the improvement is particularly pronounced for Base + CoT. This indicates that incorporating CoT, query clarification, and evaluation data positively influences the model's performance.
\textbf{2).} Models trained on Base + CoT + Clarify perform better than those trained on Base + CoT and Base + Clarify alone. This suggests that the CoT data and query clarification data complement each other for further performance improvements. Similarly, the model fine-tuned on Base + Clarify + Eval outperforms those fine-tuned on Base + Clarify and Base + Eval, indicating that the query clarification data and response evaluation data complement each other.
\textbf{3).} 
Models fine-tuned using Base$\rightarrow$Clarify outperform those fine-tuned using Base+Clarify. This is consistent with our findings from the swift study, which indicated that query clarification data is more effective when utilized in the later stages of model training.
\textbf{4).} Models fine-tuned on Base + CoT + Eval perform comparably to those fine-tuned on Base + CoT but better than the one fine-tuned on Base + Eval. This implies that CoT data encompasses most of the knowledge within the evaluation data, along with additional knowledge not present in the evaluation data. Models fine-tuned on Base + CoT + Clarify + Eval underperform those fine-tuned on Base + Clarify. 
We postulate that this is due to the inclusion of evaluation data diluting the proportion of CoT data, consequently weakening its impact. This implies that the introduction of evaluation data may not be necessary when CoT data has already been incorporated.

\begin{table}[t]
    \centering
\resizebox{\columnwidth}{!}{\begin{tabular}{l|c|c}
    \hline
        Training Data & CLLaMA2-7B                                & CLLaMA2-13B \\ \hline \hline
        Base                            & 52.74\%        & 65.42\%   \\  
        Base + CoT                      & 59.45\%        & 68.16\%   \\ 
        Base + Clarify                  & 54.62\%        & 66.03\%   \\ 
        Base $\rightarrow$ Clarify      & 56.36\%        & 67.18\%  \\
        Base+Eval                       & 55.98\%        & 66.42\%  \\ \hline
        Base+CoT+Clarify                & 62.19\%        & 69.37\% \\ 
        Base+CoT+Eval                   & 59.96\%        & 66.67\% \\
        Base+Clarify+Eval               & 57.97\%        & 66.19\% \\
        Base+CoT+Clarify+Eval           & 59.46\%        & 65.92\%  \\ \hline \hline
    \end{tabular}
}
    \caption{Model performance using training datasets built on the base dataset. Base+Clarify means training on the random mixture of the base dataset and the query clarification data. While Base$\rightarrow$Clarify means training first on the base dataset and then on the query clarification data in each epoch.}
    \label{tab:cross-task-verify}
    \vspace{-10pt}
    
\end{table}

\section{Conclusion \& Future Work}

This work introduces a gradient-based approach to assess the effectiveness of various data creation strategies for aligning Large Language Models (LLMs). Notably simple and efficient, this method requires just a handful of examples from the strategy under investigation. We employ this approach to examine the influence of Chain-of-Thought (CoT) data, query clarification data, and response evaluation data on general question-answering tasks, showcasing their remarkable ability to generalize across tasks.
To further validate our study and confirm the efficacy of our proposed methodology, we crafted training instances specific to each strategy and trained LLMs using this curated dataset. The results of this validation study align with the findings from the study using the proposed method, thereby affirming the soundness of our approach.

Although we have applied the proposed method to three intuitive strategies, we argue that it can be readily adapted to other strategies that developers may wish to explore. For instance:
\textbf{1.} We could investigate which practice is more effective: presenting the conclusion prior to the reason or the reverse, when the query necessitates a rationale but does not specify the order of reason and conclusion.
\textbf{2.} We could evaluate the potential improvement in model performance by incorporating query parsing data.
\textbf{3.} We could assess whether refining responses through multi-turn interactions can boost single-turn performance, possibly obviating the need for manual creation of single-turn training data.
\textbf{4.} We could explore which prompt template yields the greatest benefits for particular tasks.

\section{Limitations}

Our validation study primarily confirms the utility of Chain-of-Thought (CoT), query clarification, and response evaluation data. Certain additional insights derived from the swift studies await rigorous verification, such as the limited cross-task generalization capacity of non-CoT training data, the ineffective reduction of negative responses using query clarification training data, and the enhancement of some negative response probabilities even with positive training data. Furthermore, constrained by resource limitations, our experiments were conducted solely on 7B and 13B models. Exploring larger Language Models (LLMs) may uncover additional scenarios. Lastly, given our primary focus on responses in non-CoT mode, we did not conduct validation studies on public evaluation benchmarks. This decision may limit the persuasiveness of our validation study for reviewers who do not have access to our evaluation dataset.

\bibliography{acl}

\begin{thebibliography}{27}
\providecommand{\natexlab}[1]{#1}

\bibitem[{Achiam et~al.(2023)Achiam, Adler, Agarwal, Ahmad, Akkaya, Aleman, Almeida, Altenschmidt, Altman, Anadkat et~al.}]{achiam2023gpt}
Josh Achiam, Steven Adler, Sandhini Agarwal, Lama Ahmad, Ilge Akkaya, Florencia~Leoni Aleman, Diogo Almeida, Janko Altenschmidt, Sam Altman, Shyamal Anadkat, et~al. 2023.
\newblock Gpt-4 technical report.
\newblock \emph{arXiv preprint arXiv:2303.08774}.

\bibitem[{Baidu(2023)}]{wenxinyiyan}
Baidu. 2023.
\newblock 文心一言.
\newblock https://yiyan.baidu.com/.

\bibitem[{Cui et~al.(2023)Cui, Yang, and Yao}]{cui2023efficient}
Yiming Cui, Ziqing Yang, and Xin Yao. 2023.
\newblock Efficient and effective text encoding for chinese llama and alpaca.
\newblock \emph{arXiv preprint arXiv:2304.08177}.

\bibitem[{Ding et~al.(2024)Ding, Xi, He, Li, Zhai, Shi, Cai, Gui, Zhang, and Huang}]{ding2024mitigating}
Yiwen Ding, Zhiheng Xi, Wei He, Zhuoyuan Li, Yitao Zhai, Xiaowei Shi, Xunliang Cai, Tao Gui, Qi~Zhang, and Xuanjing Huang. 2024.
\newblock Mitigating tail narrowing in llm self-improvement via socratic-guided sampling.
\newblock \emph{arXiv e-prints}, pages arXiv--2411.

\bibitem[{Ekin(2023)}]{ekin2023prompt}
Sabit Ekin. 2023.
\newblock Prompt engineering for chatgpt: a quick guide to techniques, tips, and best practices.
\newblock \emph{Authorea Preprints}.

\bibitem[{Engstrom()}]{engstromdsdm}
Logan Engstrom.
\newblock Dsdm: Model-aware dataset selection with datamodels.
\newblock In \emph{Forty-first International Conference on Machine Learning}.

\bibitem[{Feldman and Zhang(2020)}]{feldman2020neural}
Vitaly Feldman and Chiyuan Zhang. 2020.
\newblock What neural networks memorize and why: Discovering the long tail via influence estimation.
\newblock \emph{Advances in Neural Information Processing Systems}, 33:2881--2891.

\bibitem[{Han et~al.(2023)Han, Simig, Mihaylov, Tsvetkov, Celikyilmaz, and Wang}]{han2023understanding}
Xiaochuang Han, Daniel Simig, Todor Mihaylov, Yulia Tsvetkov, Asli Celikyilmaz, and Tianlu Wang. 2023.
\newblock Understanding in-context learning via supportive pretraining data.
\newblock In \emph{The 61st Annual Meeting Of The Association For Computational Linguistics}.

\bibitem[{Hu et~al.(2021)Hu, Shen, Wallis, Allen-Zhu, Li, Wang, Wang, and Chen}]{hu2021lora}
Edward~J Hu, Yelong Shen, Phillip Wallis, Zeyuan Allen-Zhu, Yuanzhi Li, Shean Wang, Lu~Wang, and Weizhu Chen. 2021.
\newblock Lora: Low-rank adaptation of large language models.
\newblock \emph{arXiv preprint arXiv:2106.09685}.

\bibitem[{Liu et~al.()Liu, Zeng, He, Jiang, and He}]{liumakes}
Wei Liu, Weihao Zeng, Keqing He, Yong Jiang, and Junxian He.
\newblock What makes good data for alignment? a comprehensive study of automatic data selection in instruction tuning.
\newblock In \emph{The Twelfth International Conference on Learning Representations}.

\bibitem[{Liu et~al.(2024)Liu, Zeng, He, Jiang, and He}]{liu2024what}
Wei Liu, Weihao Zeng, Keqing He, Yong Jiang, and Junxian He. 2024.
\newblock \href {https://openreview.net/forum?id=BTKAeLqLMw} {What makes good data for alignment? a comprehensive study of automatic data selection in instruction tuning}.
\newblock In \emph{The Twelfth International Conference on Learning Representations}.

\bibitem[{Madsen et~al.(2022)Madsen, Reddy, and Chandar}]{madsen2022post}
Andreas Madsen, Siva Reddy, and Sarath Chandar. 2022.
\newblock Post-hoc interpretability for neural nlp: A survey.
\newblock \emph{ACM Computing Surveys}, 55(8):1--42.

\bibitem[{OpenAI(2022)}]{openai2022openai}
Openai OpenAI. 2022.
\newblock Openai: Introducing chatgpt.
\newblock \emph{URL https://openai. com/blog/chatgpt}.

\bibitem[{Ouyang et~al.(2022)Ouyang, Wu, Jiang, Almeida, Wainwright, Mishkin, Zhang, Agarwal, Slama, Ray et~al.}]{ouyang2022training}
Long Ouyang, Jeffrey Wu, Xu~Jiang, Diogo Almeida, Carroll Wainwright, Pamela Mishkin, Chong Zhang, Sandhini Agarwal, Katarina Slama, Alex Ray, et~al. 2022.
\newblock Training language models to follow instructions with human feedback.
\newblock \emph{Advances in neural information processing systems}, 35:27730--27744.

\bibitem[{Pruthi et~al.(2020)Pruthi, Liu, Kale, and Sundararajan}]{pruthi2020estimating}
Garima Pruthi, Frederick Liu, Satyen Kale, and Mukund Sundararajan. 2020.
\newblock Estimating training data influence by tracing gradient descent.
\newblock \emph{Advances in Neural Information Processing Systems}, 33:19920--19930.

\bibitem[{Shen(2024)}]{shen2024rethinking}
Ming Shen. 2024.
\newblock Rethinking data selection for supervised fine-tuning.
\newblock \emph{arXiv preprint arXiv:2402.06094}.

\bibitem[{Sprague et~al.(2024)Sprague, Yin, Rodriguez, Jiang, Wadhwa, Singhal, Zhao, Ye, Mahowald, and Durrett}]{sprague2024cot}
Zayne Sprague, Fangcong Yin, Juan~Diego Rodriguez, Dongwei Jiang, Manya Wadhwa, Prasann Singhal, Xinyu Zhao, Xi~Ye, Kyle Mahowald, and Greg Durrett. 2024.
\newblock To cot or not to cot? chain-of-thought helps mainly on math and symbolic reasoning.
\newblock \emph{arXiv preprint arXiv:2409.12183}.

\bibitem[{Sun et~al.(2024)Sun, Shen, Zhou, Zhang, Chen, Cox, Yang, and Gan}]{sun2024principle}
Zhiqing Sun, Yikang Shen, Qinhong Zhou, Hongxin Zhang, Zhenfang Chen, David Cox, Yiming Yang, and Chuang Gan. 2024.
\newblock Principle-driven self-alignment of language models from scratch with minimal human supervision.
\newblock \emph{Advances in Neural Information Processing Systems}, 36.

\bibitem[{Touvron et~al.(2023)Touvron, Martin, Stone, Albert, Almahairi, Babaei, Bashlykov, Batra, Bhargava, Bhosale et~al.}]{touvron2023llama}
Hugo Touvron, Louis Martin, Kevin Stone, Peter Albert, Amjad Almahairi, Yasmine Babaei, Nikolay Bashlykov, Soumya Batra, Prajjwal Bhargava, Shruti Bhosale, et~al. 2023.
\newblock Llama 2: Open foundation and fine-tuned chat models.
\newblock \emph{arXiv preprint arXiv:2307.09288}.

\bibitem[{Wang et~al.(2023{\natexlab{a}})Wang, Tan, Guo, and Li}]{wang2023noise}
Song Wang, Zhen Tan, Ruocheng Guo, and Jundong Li. 2023{\natexlab{a}}.
\newblock Noise-robust fine-tuning of pretrained language models via external guidance.
\newblock In \emph{Findings of the Association for Computational Linguistics: EMNLP 2023}, pages 12528--12540.

\bibitem[{Wang et~al.(2023{\natexlab{b}})Wang, Kordi, Mishra, Liu, Smith, Khashabi, and Hajishirzi}]{wang2023self}
Yizhong Wang, Yeganeh Kordi, Swaroop Mishra, Alisa Liu, Noah~A Smith, Daniel Khashabi, and Hannaneh Hajishirzi. 2023{\natexlab{b}}.
\newblock Self-instruct: Aligning language models with self-generated instructions.
\newblock In \emph{The 61st Annual Meeting Of The Association For Computational Linguistics}.

\bibitem[{Weber et~al.(2024)Weber, Litschko, Artemova, and Plank}]{weber2024donkii}
Leon Weber, Robert Litschko, Ekaterina Artemova, and Barbara Plank. 2024.
\newblock Donkii: Characterizing and detecting errors in instruction-tuning datasets.
\newblock In \emph{Proceedings of The 18th Linguistic Annotation Workshop (LAW-XVIII)}, pages 197--215.

\bibitem[{Wei et~al.(2022)Wei, Wang, Schuurmans, Bosma, Xia, Chi, Le, Zhou et~al.}]{wei2022chain}
Jason Wei, Xuezhi Wang, Dale Schuurmans, Maarten Bosma, Fei Xia, Ed~Chi, Quoc~V Le, Denny Zhou, et~al. 2022.
\newblock Chain-of-thought prompting elicits reasoning in large language models.
\newblock \emph{Advances in neural information processing systems}, 35:24824--24837.

\bibitem[{White et~al.(2023)White, Fu, Hays, Sandborn, Olea, Gilbert, Elnashar, Spencer-Smith, and Schmidt}]{white2023prompt}
Jules White, Quchen Fu, Sam Hays, Michael Sandborn, Carlos Olea, Henry Gilbert, Ashraf Elnashar, Jesse Spencer-Smith, and Douglas~C Schmidt. 2023.
\newblock A prompt pattern catalog to enhance prompt engineering with chatgpt.
\newblock \emph{arXiv preprint arXiv:2302.11382}.

\bibitem[{Xia et~al.(2024)Xia, Malladi, Gururangan, Arora, and Chen}]{xialess}
Mengzhou Xia, Sadhika Malladi, Suchin Gururangan, Sanjeev Arora, and Danqi Chen. 2024.
\newblock Less: Selecting influential data for targeted instruction tuning.
\newblock In \emph{Forty-first International Conference on Machine Learning}.

\bibitem[{Zhang et~al.(2023)Zhang, Dong, Li, Zhang, Sun, Wang, Li, Hu, Zhang, Wu et~al.}]{zhang2023instruction}
Shengyu Zhang, Linfeng Dong, Xiaoya Li, Sen Zhang, Xiaofei Sun, Shuhe Wang, Jiwei Li, Runyi Hu, Tianwei Zhang, Fei Wu, et~al. 2023.
\newblock Instruction tuning for large language models: A survey.
\newblock \emph{arXiv preprint arXiv:2308.10792}.

\bibitem[{Zhou et~al.(2024)Zhou, Liu, Xu, Iyer, Sun, Mao, Ma, Efrat, Yu, Yu et~al.}]{zhou2024lima}
Chunting Zhou, Pengfei Liu, Puxin Xu, Srinivasan Iyer, Jiao Sun, Yuning Mao, Xuezhe Ma, Avia Efrat, Ping Yu, Lili Yu, et~al. 2024.
\newblock Lima: Less is more for alignment.
\newblock \emph{Advances in Neural Information Processing Systems}, 36.

\end{thebibliography}

\newpage

\appendix

\section{Data Creation using the Study Strategies}
Table \ref{tab:probe_case} reveals the process of the studied strategies for generating probe examples from reference examples. 

Let $(q, r)$ represent a reference example. For creating a corresponding CoT training example for both the swift study and validation study, we appended a directive, ``Please output the reason first and then the final answer, regardless of the above description.", to $q$, creating a modified query, $q'$. Following this, we prompted Yiyan to generate a CoT response, $r'$, for $q'$ referencing $r$ to ensure consistency between $r$ and $r$. The pair $(q', r')$ thus forms the corresponding CoT instance of $(q, r)$. A similar process is applied to generate the response evaluation example but with a different prompt.

For creating a corresponding query clarification training example for the swift study, we manually revised $q$ to obtain the revised query. While in the validation study, we prompt Yiyan to extract the requirements in $q$. The extracted requirements were then appended to $q$, creating clarified query, $q'$. The pairing of $q$ and $q'$ formed a query clarification example.

\begin{table*}[]
    \centering
    \begin{tabular}{l|l} 
        \toprule 
        Strategy & Generated Example \\ 
        \midrule 
        CoT 
        
        & \pbox{13cm}{\textbf{Query}: Are the following two sentences semantically similar: 1. xxx 2. xxx. Answer directly with similar/not similar. \textit{Please output the reason first and then the final answer, regardless of the above description.}\\ 
        \textbf{Response}: Sentence 1 talks about xxx. Sentence 2 talks about xxx. They xxx. So the answer is xxx.} \\ 
        \midrule
        \pbox{1.8cm}{Query \\ Clarification}
        
        & \pbox{13cm}{\textbf{Query}: [Input]: `With the first word’s initial as K and the second word’s initial as S, compose a phrase.'
        \textit{It is known that the [input] is a user's question, which may not be clearly expressed. Please help me rephrase this question to make its meaning clearer.}  \\
        \textbf{Response}: Form a phrase with the initial letter of the first word as K and the initial letter of the second word as S.} \\
        \midrule
         
        \pbox{1.5cm}{Response \\ Evaluation}
        
        & \pbox{13cm}{\textbf{Query}: \textit{It is known that the [input] stores the questions posed by the user to the dialogue system, and the [output] stores the responses of the dialogue system to the [input]. Please help me judge whether the following [output] appropriately responds to the [input].} 
        [Input]: xxx
        [Output]: xxx
        \\
        \textbf{Response}: First, let's list the requirements of the input: xxx. Next, we will analyze the output based on these requirements: xxx.
        Thus, the answer is: the output meets/does not meet all the requirements of the input.} \\
        \midrule
        \bottomrule 
    \end{tabular}
    \caption{Generation of examples using the three studied strategies. Each example is derived from a reference example, with the words in italic font serving as the prompt for wrapping the reference example.}
    \label{tab:probe_case}
\end{table*}

\section{Reference Examples for Swift Studies}
Table \ref{tab:reference_case} shows some reference examples used for generating probe examples in the swift studies. 

\begin{table*}[t]
    \centering
    \begin{tabular}{l|l|l} 
        \toprule 
        ID & Query & Task \\ 
        \midrule 
        1 & \pbox{9.5cm}{Xiao Zhang bought 5 pens, and Xiao Li bought 2 books. They exchanged their items. How many books does Xiao Zhang have now? Just give the answer.} & Logic Reasoning \\ 
        \midrule
        2 & \pbox{9.5cm}{Given a label set: xxx, xxx, determine the topic category of the following paragraph and select the most appropriate tag: xxx.} & Classification \\
        \midrule
        3 & Convert 3,894,544 yuan to its uppercase form. & Chinese Digit Conversion \\ 
        \midrule
        4 & \pbox{9.5cm}{Are the following two sentences semantically similar: 1. xxx 2. xxx. Answer directly with similar/not similar.} & Semantic Similarity \\ 
        \midrule
        5 & \pbox{9.5cm}{Identify any typographical errors in the following paragraph: xxx.} & Error Correction \\ 
        \midrule
        6 & \pbox{9.5cm}{Form a word with the first character's pinyin initial as T and the second character's pinyin initial as H.} & Chinese Word Formation \\ 
        \midrule
        7 & \pbox{9.5cm}{Given the following action labels: xxx, xxx, please insert a reasonable action label after each short sentence in the input text: xxx.} & Controlled Text Edit \\
        \midrule
        8 & \pbox{9.5cm}{Evaluate the quality of answers. All inputs and outputs are in json format. [Input] xxx (a json). Please provide the scores for answer A and answer B in the output, and list the deduction points in the reasons. The following are special requirements to pay attention to: xxx. [Output Format]: xxx.} & Paragraph Analysis \\ 
        \midrule
        9 & Read the following article and provide a brief summary: xxx. & Summarization \\
        \midrule
        10 & \pbox{9.5cm}{Based on the given historical question and rewritten question, generate a new question. xxx. Historical question: xxx Rewritten question: xxx Result requirements: xxx, Current time: xxx.} & Controlled Question Generation \\ \midrule
        11 & \pbox{9.5cm}{Please divide the following text into 3 paragraphs based on its meaning, ensuring clear distinctions between paragraphs and compactness within each paragraph. Add paragraph numbers at the beginning of each paragraph (e.g., "1. ") and an end marker at the end of each paragraph (e.g., "——END——").
Input: xxx} & Text Editing\\ \midrule
        12 & \pbox{9.5cm}{In the following text, there is information about a character named A. xxx. Please extract and organize all the information about A, present it in a list format, and provide the corresponding attribute names.} & Information Extraction \\
        \bottomrule 
    \end{tabular}
    \caption{Reference examples used in the swift study from distinct tasks.}
    \label{tab:reference_case}
\end{table*}

\end{CJK*}
\end{document}